\documentclass{article}
\usepackage{iclr2023_conference,times}

\usepackage{amsmath,amsfonts,bm}

\def\eqref#1{equation~\ref{#1}}

\def\1{\bm{1}}

\DeclareMathAlphabet{\mathsfit}{\encodingdefault}{\sfdefault}{m}{sl}
\SetMathAlphabet{\mathsfit}{bold}{\encodingdefault}{\sfdefault}{bx}{n}

\usepackage{graphicx}

\usepackage{array}
\usepackage{ifthen}
\usepackage[normalem]{ulem}
\newcommand{\niv}[2][]{%
    \ifthenelse{ \equal{#1}{} }
        {\textcolor{red}{(NH) #2}}
        {\textcolor{red}{(NH) \sout{#1} #2}}
}
\usepackage{wrapfig}
\usepackage{caption}

\newcommand{\bx}{\mathbf{x}}
\newcommand{\btheta}{{\boldsymbol{\theta}}}
\newcommand{\reals}{{\mathbb R}}
\newcommand{\norm}[1]{\left\|#1\right\|}
\newcommand{\zero}{{\mathbf{0}}}

\usepackage{hyperref}
\usepackage{url}

\usepackage{cleveref}

\title{Reconstructing Training Data from \\Multiclass Neural Networks}

\newcommand*{\affaddr}[1]{#1}
\newcommand*{\affmark}[1][*]{\textsuperscript{#1}}
\newcommand*{\affmarkt}[1][*]{\textsuperscript{~~~#1}}

\author{%
Gon Buzaglo\thanks{Equal Contribution}\affmarkt[1]~,~ Niv Haim\footnotemark[1]\affmarkt[1]~,~ Gilad Yehudai\affmark[1], ~Gal Vardi\affmark[2] \& Michal Irani\affmark[1]\\
\affaddr{\affmark[1]Weizmann Institute of Science, Rehovot, Israel}\\
\affaddr{\affmark[2]TTI Chicago and the Hebrew University of Jerusalem}
}

\newdimen\mywidth

\iclrfinalcopy 
\begin{document}

  \mywidth=\dimexpr
    \linewidth
    - 20\tabcolsep
  \relax

\maketitle

\begin{figure}[ht]
\centering
\begin{tabular}{%
    p{.1\mywidth}%
    p{.1\mywidth}%
    p{.1\mywidth}%
    p{.1\mywidth}%
    p{.1\mywidth}%
    p{.1\mywidth}%
    p{.1\mywidth}%
    p{.1\mywidth}%
    p{.1\mywidth}%
    p{.1\mywidth}%
}

\hfill Plane & \hfill Car & \hfill Bird & \hfill Cat & \hfill Deer & \hfill Dog & \hfill Frog & \hfill Horse & \hfill Ship & \hfill Truck \\
     \multicolumn{10}{l}{\includegraphics[width=\linewidth]{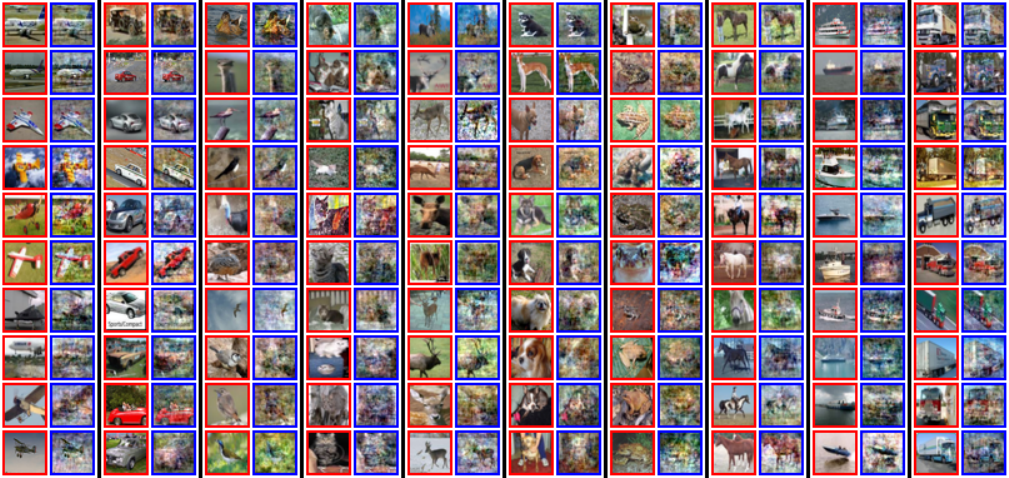}} 
\end{tabular}
\caption{Reconstructed training samples from a multi-class MLP classifier that was trained on $500$ CIFAR10 images. Each column corresponds to one class and shows the $10$ training samples (\textcolor{red}{\textit{red}}) that were best reconstructed from this class, along with their reconstructed result (\textcolor{blue}{\textit{blue}}).}
\label{fig:margins_res}
\end{figure}

\begin{abstract}
Reconstructing samples from the training set of trained neural networks is a major privacy concern. \cite{haim2022reconstructing} recently showed that it is possible to reconstruct training samples from neural network binary classifiers, based on theoretical results about the implicit bias of gradient methods.
In this work, we present several improvements and new insights over this previous work. 
As our main improvement, we
show that training-data reconstruction is possible in the multi-class setting and that the reconstruction quality is even higher than in the case of binary classification.  

Moreover, we show that using weight-decay during training increases the vulnerability to sample reconstruction. Finally, while in the previous work the training set was of size at most $1000$ from $10$ classes, we show preliminary evidence of the ability to reconstruct from a model trained on $5000$ samples from $100$ classes.
\end{abstract}

\section{Introduction}

Understanding memorization in data-driven machine learning models is a fundamental question with implications on explainability, privacy, artistic synthesis and more. ~\cite{haim2022reconstructing} recently demonstrated that a large portion of the training samples are encoded in the parameters of trained neural network binary classifiers, by explicitly reconstructing samples from the training set of such models. 
The reconstruction method is based on consequences of the implicit bias of gradient descent, as presented by~\cite{lyu2019gradient,ji2020directional} -- a homogeneous neural network trained with gradient descent will converge (in direction) to a solution of the KKT conditions of a maximum-margin problem. This dictates a set of equations that relates the parameters of the trained network and the training data. The key observation is that given a trained model (and its parameters), these relations can be leveraged to reconstruct training samples. Thus, a loss function is devised to show that by changing the inputs to the classifier (in order to minimize the loss), the inputs converge to true samples from the original training set.

The work of \cite{haim2022reconstructing} had several limitations, namely: (1) Their work only showed reconstructions from binary classifiers; (2) For their reconstruction method to succeed the trained network needed to be initialized with very small weights, smaller than standard Xavier initialization \citep{glorot2010understanding}; and (3) Their experiments consisted of only small datasets with at most $1000$ samples. In this work we extend their results in several directions, and overcome some of the limitations while gaining new insights on this reconstruction method.

\paragraph{Our contributions:}(1) We show reconstruction of large portions of actual training samples from a trained \emph{multi-class} neural networks; (2) We show that the use of weight-decay enables reconstruction of samples from models trained with standard initialization schemes thus overcoming a major limitation of~\cite{haim2022reconstructing}; (3) We show reconstruction of models trained on datasets that are $10$x larger than shown in~\cite{haim2022reconstructing}; (4) We empirically analyze the effect of weight-decay on sample reconstruction, showing that it increases the vulnerability to such attacks.

\paragraph{Related works.} Several works have shown different privacy attacks in deep learning architectures, which aim to leak private information from trained models. For example, \emph{Model inversion} attacks aim at reconstructing class representatives \cite{fredrikson2015model, he2015delving, yang2019neural}.
Other types of attacks target specific models, such as extracting data from \emph{language models} \cite{carlini2019secret,carlini2021extracting}, which use crafted prompts; and information leakage from collaborative deep learning (federated learning) \cite{he2019model, melis2019exploiting,huang2021evaluating,hitaj2017deep}. In \cite{balle2022reconstructing} a reconstruction attack is shown where the attacker knows all the training samples except for one.
Recently, \cite{carlini2023extracting} showed extraction of actual training images from trained diffusion models. Their methods rely on generating many different images and using known membership inference attacks to determine which generated image was used as a training sample. We emphasize that their method is specific for generative models, whereas we focus on classifiers.

\section{Preliminaries - Implicit Bias of Gradient Methods}

Neural networks are commonly trained using gradient methods, and when large enough, they are expected to fit the training data well. However, it is empirically known that these models converge to solutions that also generalize well to unseen data, despite the risk of overfitting. Several works pointed to the ``\emph{implicit bias}" of gradient methods as a possible explanation. One of the most prominent results in this area is by \cite{soudry2018implicit}, who showed that linear classifiers
trained with gradient descent on the logistic loss converge to the same solution as that of a hard-SVM, meaning that they maximize the margins. This result was later extended to non-linear and homogeneous neural networks by \cite{lyu2019gradient,ji2020directional}. Based on these results, \cite{haim2022reconstructing} have devised a data reconstruction scheme from trained binary classifiers (see Section 3 in \cite{haim2022reconstructing}).
Below we describe an extension of the theorem about the implicit bias of homogeneous neural networks to a multi-class setup, based on theoretical results from Appendix G in \cite{lyu2019gradient}.

Formally, let $S = \{(\bx_i,y_i)\}_{i=1}^n \subseteq \reals^d \times [C]$ be a multi-class classification training set where $C\in\mathbb{N}$ is any number of classes, and $[C]= \{1,\dots,C\}$. Let $\Phi(\btheta;\cdot):\reals^d \to \reals^C$ be a neural network parameterized by $\btheta \in \reals^p$. We denote the $j$-th output of $\Phi$ on an input $\bx$ as $\Phi_j(\btheta;\bx) \in \reals$. Consider a homogeneous network, minimizing the standard cross-entropy loss and assume that after some number of iterations the model correctly classifies all the training examples. Then, gradient flow will converge to a KKT point of the following maximum-margin problem:

\begin{equation}
\label{eq:optimization problem}
	\min_{\btheta} \frac{1}{2} \norm{\btheta}^2 \;\;\;\; \text{s.t. } \;\;\; \Phi_{y_i}(\btheta; \bx_i) - \Phi_{j}(\btheta; \bx_i) \geq 1 ~~~ \forall i \in [n], \forall j \in [C] \setminus \{y_i\} \;\; ~.
\end{equation}

This KKT point is characterized by the following set of conditions:
\begin{align}
    &\btheta - \sum_{i=1}^n \sum_{j\ne{y_i}}^c\lambda_{i,j}  \nabla_{\btheta}  ( \Phi_{y_i}(\btheta; \bx_i) - \Phi_{j}(\btheta; \bx_i) ) = \zero~ \label{eq:stationary}\\
    &\forall i \in [n], \forall j \in [C] \setminus \{y_i\}: \;\;  \Phi_{y_i}(\btheta; \bx_i) - \Phi_{j}(\btheta; \bx_i) \geq 1 \label{eq:prim feas} \\
    &\forall i \in [n], \forall j \in [C] \setminus \{y_i\}: \;\;  \lambda_{i,j} \geq 0 \label{eq:dual feas}\\
    &\forall i \in [n], \forall j \in [C] \setminus \{y_i\}: \;\;  \lambda_{i,j}= 0 ~ \text{if}~  \Phi_{y_i}(\btheta; \bx_i) - \Phi_{j}(\btheta; \bx_i) \neq 1 \label{eq:comp slack}
\end{align}

\section{Multi-Class Reconstruction}

We use a similar reconstruction method to \cite{haim2022reconstructing}. Suppose we are given a trained classifier with parameters $\btheta$, our goal is to find the set of data samples $\{\bx_i\}_{i=1}^n$ that the network trained on.
A straightforward approach for such a loss would be to minimize the norm of the L.H.S of condition~\cref{eq:stationary}. That is, we initialize  $\{\bx_i\}_{i=1}^m$ and $\{\lambda_{i,j}\}_{i \in [n], j\in [C] \setminus y_i}$ where $m$ is a hyperparameter.

Note that from \cref{eq:prim feas,eq:comp slack}, most $\lambda_{i,j}$ zero out:  the distance of a sample $\bx_i$ to its nearest decision boundary, \mbox{$\Phi_{y_i} - \max_{j\ne{y_i}}\Phi_j$}, is usually achieved for a single class $j$ and therefore (from \cref{eq:comp slack}) in this case at most one $\lambda_{i,j}$ will be non-zero. For some samples $\bx_i$ it is also possible that all $\lambda_{i,j}$ will vanish.
We therefore turn to only optimizing on the distance from the decision boundary. This implicitly includes~\cref{eq:comp slack} into the summation in~\cref{eq:stationary}, dramatically reducing the number of summands, and simplifying the overall optimization problem. We define the following loss:

\begin{equation}
    L_{st}(\bx_1,...,\bx_m,\lambda_1,...,\lambda_m)=\left\| \btheta - \sum_{i=1}^{m}\lambda_i\ \nabla_\btheta[\Phi_{y_i}(\bx_i;\btheta)-\max_{j\ne{y_i}}\Phi_j(\bx_i;\btheta)]\right\|_2^2
\label{eq:simplified_loss}
\end{equation}

Our reconstruction method is, given the parameters of a trained network $\btheta$, initialize $\bx_i$ and $\lambda_i$ for $i=1,\dots,m$, and minimize \eqref{eq:simplified_loss}.
In order to make sure that \cref{eq:dual feas} is satisfied we optimize for $a_i$ and require that $\lambda_i=a_i^2$\footnote{More precisely, $\lambda_i=a_i^2 + \lambda_\text{min}$, where $\lambda_\text{min}$ is a hyperparameter that encourages the reconstructed samples to converge to margin-samples.}. This further simplifies the optimization problem compared to \cite{haim2022reconstructing} that use a separate loss function.

Since $n$ is unknown we set $m \geq n$ which represents the number of samples we want to reconstruct (thus, we only need to upper bound $n$). We can hypothetically set $m= C\cdot n$ and with balanced labels, this way there are enough reconstructed samples for any distribution of the labels. In practice, we set $m$ to be slightly larger than $n$, which is enough to get good reconstructions. The rest of the hyperparameters (including $\lambda_{\min}$) are chosen using a hyper-parameter search. 

\section{Results}

\subsection{Multiclass Reconstruction}

We compare between reconstruction from binary classifiers (as studied in \cite{haim2022reconstructing}) and multi-class classifiers. We conduct the following experiment: we train an MLP classifier with architecture \mbox{$D$-$1000$-$1000$-$C$} on $500$ samples from the CIFAR10 \citep{krizhevsky2009learning} dataset. We use full-batch GD, and set the learning rate as $0.5$. 
The model is trained to minimize the cross-entropy loss with full-batch gradient descent, once with two classes ($250$ samples per class) and once for the full $10$ classes ($50$ samples per class).
The test set accuracy of the models is $77\%$/$32\%$ respectively, which is far from random ($50\%$/$10\%$ resp.).

\begin{figure}[htbp]
    \includegraphics[width=\textwidth]{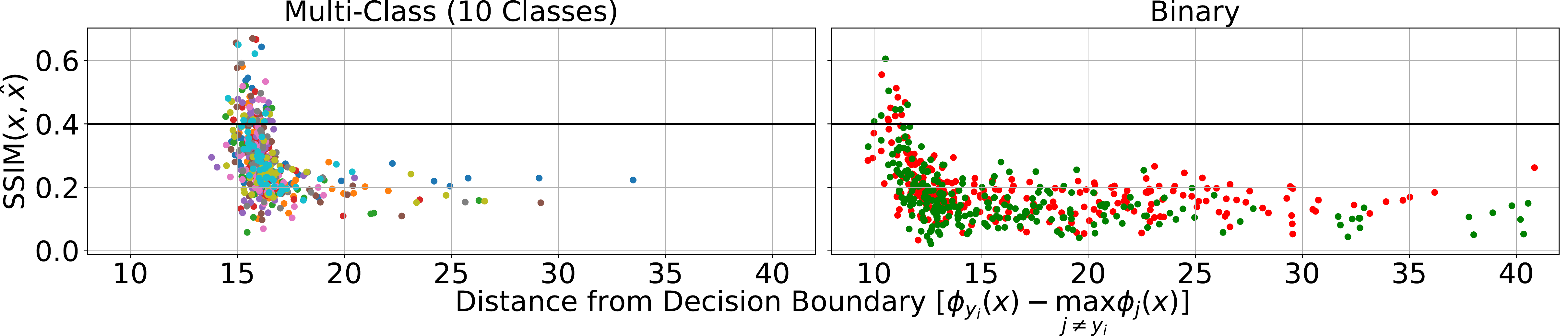}
    \caption{Multi-class classifiers are more vulnerable to training-set reconstruction. For a training set of size $500$, a multi-class model (\textit{left}) yields much more reconstructed samples with good quality (SSIM$>0.4$), than a binary classification model (\textit{right}).}
    \label{fig:binary_vs_multiclass}
    \centering
\end{figure}

To quantify the quality of our reconstructed samples, for each sample in the original training set we search for its nearest neighbour in the reconstructed images and measure the similarity using SSIM~\citep{wang2004image} (higher SSIM means better reconstruction). As a rule of thumb, we say that a sample was reconstructed well if its SSIM$>0.4$ (see discussion in \cref{sec:ssim_0.4}). In \cref{fig:binary_vs_multiclass} we plot the quality of reconstruction (in terms of SSIM) against the distance of the sample from the decision boundary $\Phi_{y_i}(\bx_i;\btheta)-\max_{j\ne{y_i}}\Phi_j(\bx_i;\btheta)$.  As seen, a multi-class classifier yields much more samples that are vulnerable to being reconstructed.

\begin{figure}[!hb]
    \begin{tabular}{c}
         \includegraphics[width=\textwidth]{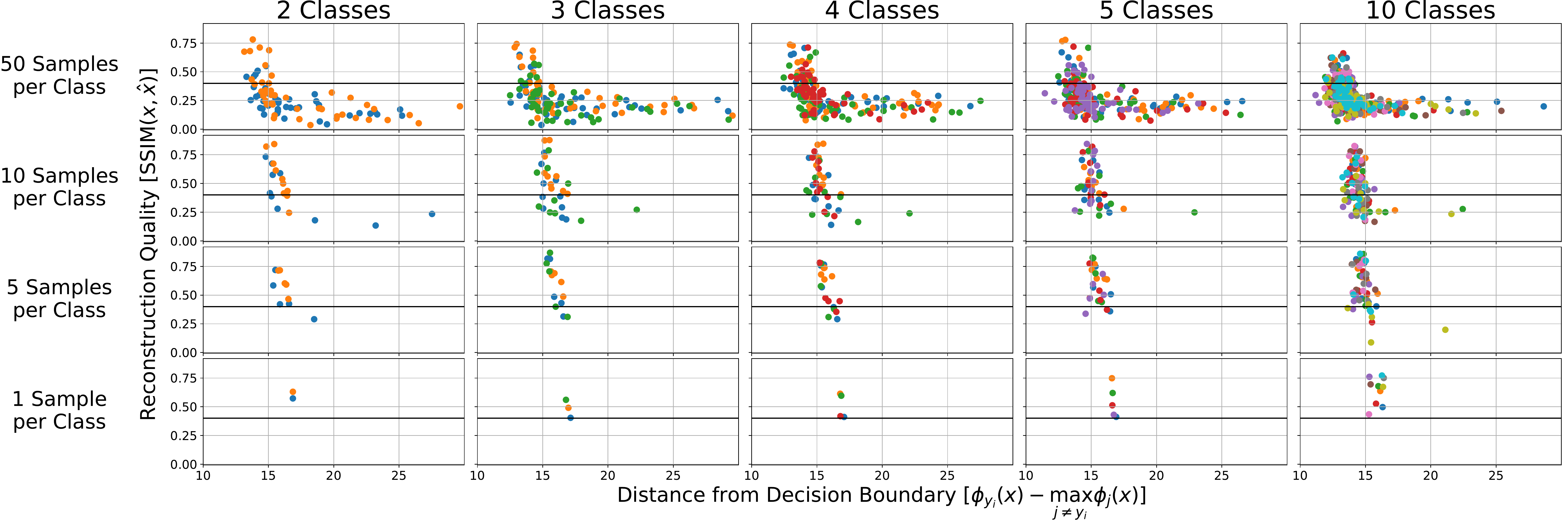}  \\
         (a) Full results of each experiment \\
         \includegraphics[width=\textwidth]{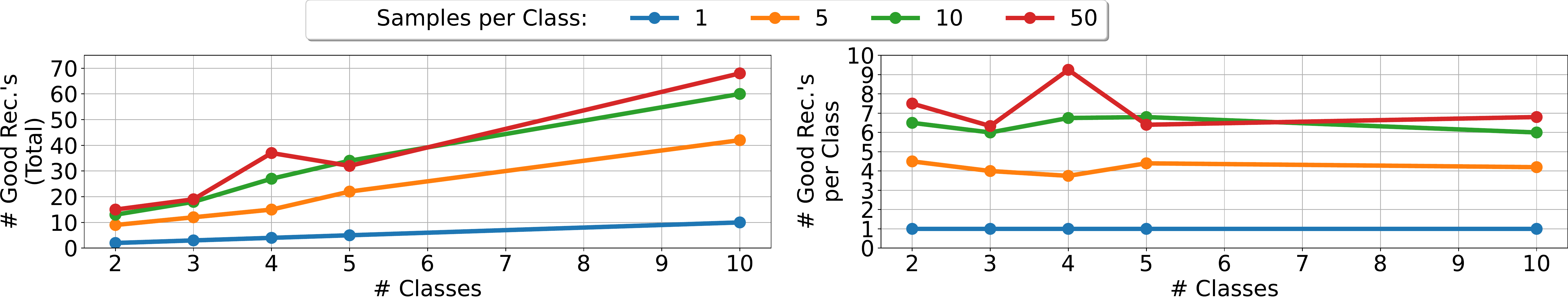} \\
         (b) Number of ``good" reconstructions increases with number 
         of classes and the samples per class
    \end{tabular}
    \caption{Evaluating effect of multiple classes on the ability to reconstruct. We show reconstructions from models trained with different number of classes and different number of samples per class. As seen, multiple classes result in more reconstructed samples.}
    \label{fig:c_vs_dpc}
    \centering
\end{figure}

Next, we examine the dependence between the ability to reconstruct from a model and the number of classes on which it was trained. Comparing between two models trained on different number of classes is not immediately clear, since we want to isolate the effect of the number of classes from the size of the dataset (it was observed by \citet{haim2022reconstructing} that the number of reconstructed samples decreases as the total size of the training set increases). We therefore train models on training sets with varying number of classes ($C \in \{2,3,4,5,10\}$) and varying number of samples per class ($1,5,10,50$). The results are visualized in \cref{fig:c_vs_dpc}a. Note that for models with same number of samples per class, the ability to reconstruct \emph{increases} with the number of classes, even though the total size of the training set is larger. This further validates our hypothesis that multi-class models are more vulnerable to reconstruction. We continue this study in \cref{sec:fixed_dataset_size}.

Another way to validate this hypothesis is by showing the dependency between the number of classes and the number of ``good" reconstructions (SSIM$>0.4$) -- shown in~\cref{fig:c_vs_dpc}b. As can be seen, training on multiple classes yields more samples that are vulnerable to reconstruction.
A possible intuitive explanation, is that multi-class classifiers have more ``margin" samples. Since margin-samples are more vulnerable to reconstruction, this results in more samples being reconstructed from the model.

\subsection{Weight Decay Increases Reconstructability}
\noindent%
\begin{minipage}{\linewidth}
\makebox[\linewidth]{
    \begin{tabular}{cc}
         \includegraphics[keepaspectratio=true,width=.45\textwidth]{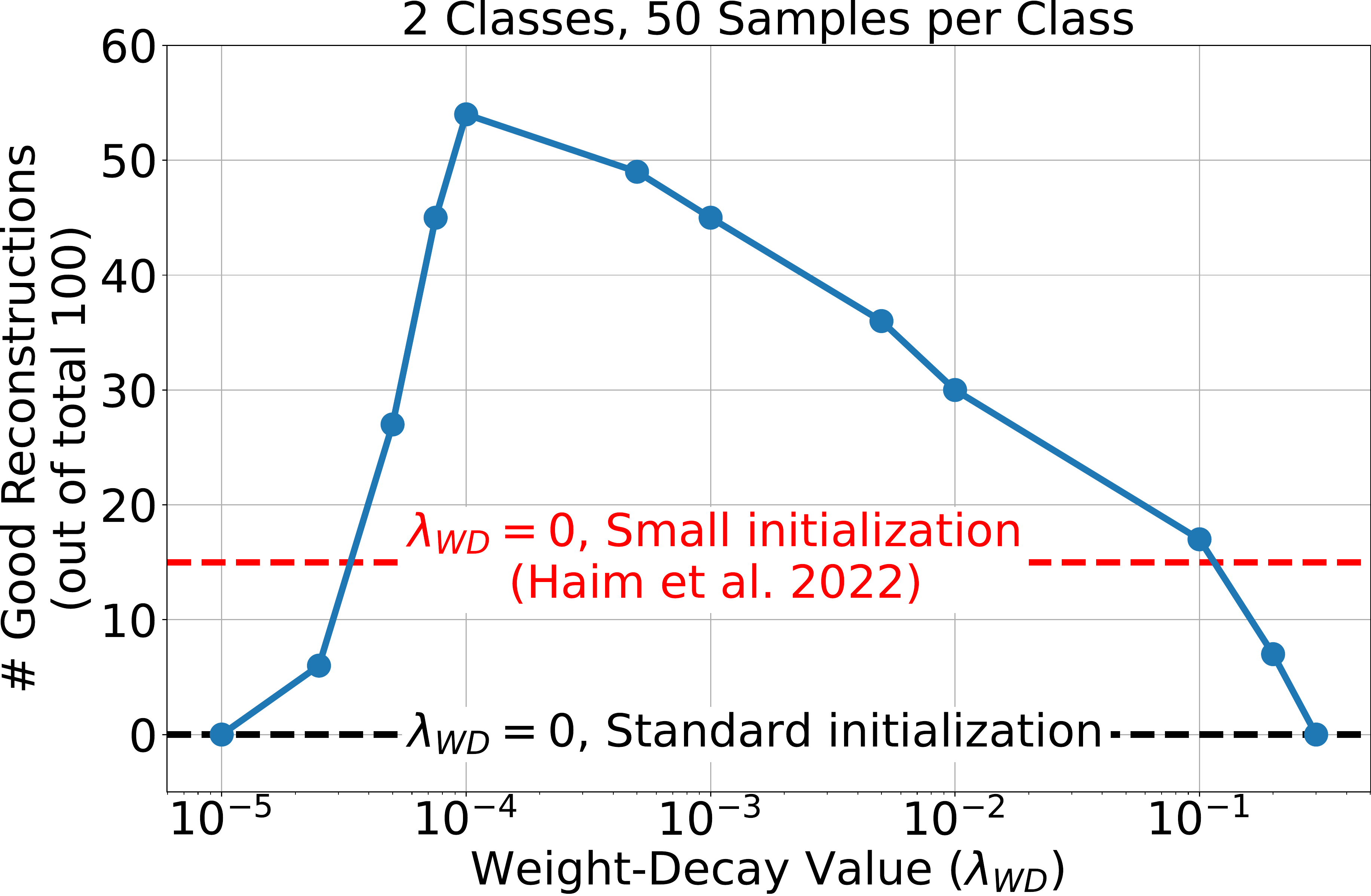} &  
         \includegraphics[keepaspectratio=true,width=.45\textwidth]{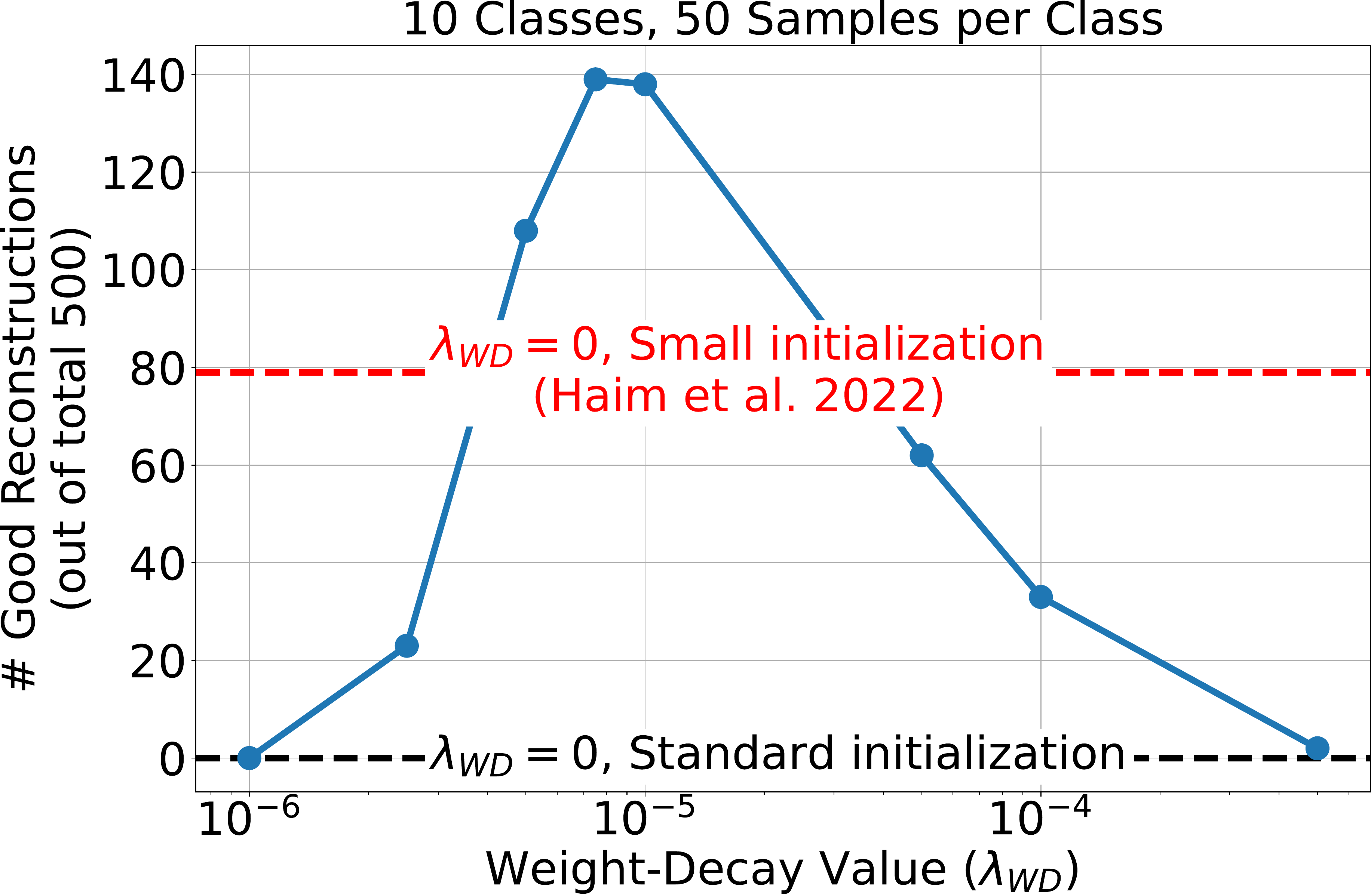} \\
    \end{tabular}
  }
\captionof{figure}{Using weight-decay during training increases vulnerability to sample reconstruction}\label{fig:weight_decay} 
\end{minipage}

Another drawback of \cite{haim2022reconstructing} is that reconstruction was only shown for models whose first fully-connected layer was initialized with small (non-standard) weights. Models with standard initialization, such as Kaiming \cite{he2015delving} or Xavier \cite{glorot2010understanding} where each weight vector is initialized with a variance of $\sim\frac{1}{d}$ (where $d$ is the input's layer dimension) did not yield good reconstructed samples. \cite{haim2022reconstructing} only reconstructed from networks initialized with a variance of $\sim\frac{1}{d^{1.5}}$. Set to better understand this drawback, we observed an interesting phenomenon -- for models with standard initialization, using weight-decay during training enabled samples reconstruction. Moreover, for some values of weight-decay the reconstructability is \emph{significantly higher} than what was observed for models with small-initialized-first-layer models. In \cref{fig:weight_decay} we show the number of good reconstructed samples for different choices of the value of the weight decay ($\lambda_{\text{WD}}$). We show results for two models trained on $C=2$ classes (\cref{fig:weight_decay}\textit{left}) and $C=10$ classes (\cref{fig:weight_decay}\textit{right}), both trained on $50$ samples per class. We add two baselines trained without weight-decay: model trained with standard initialization (\textit{black}) and model with small-initialized-first-layer (\textit{red}).

By looking at the exact distribution of reconstruction quality to the distance from the margin, we observe that weight-decay (for some values) results in more training samples being on the margin of the trained classifier, thus being more vulnerable to reconstruction. This observation is shown in~\cref{fig:weight_decay_scatter} where we show the plots for all experiments from~\cref{fig:weight_decay}\textit{left}. We also provide the train and test errors for each model. It seems that the generalization (test error) does not change significantly. However, an interesting observation is that reconstruction is possible even for models with non-zero training errors (for which, the assumptions of~\cite{lyu2019gradient} do not hold).

\noindent%
\begin{minipage}{\linewidth}
\makebox[\linewidth]{
  \includegraphics[keepaspectratio=true,width=\textwidth]{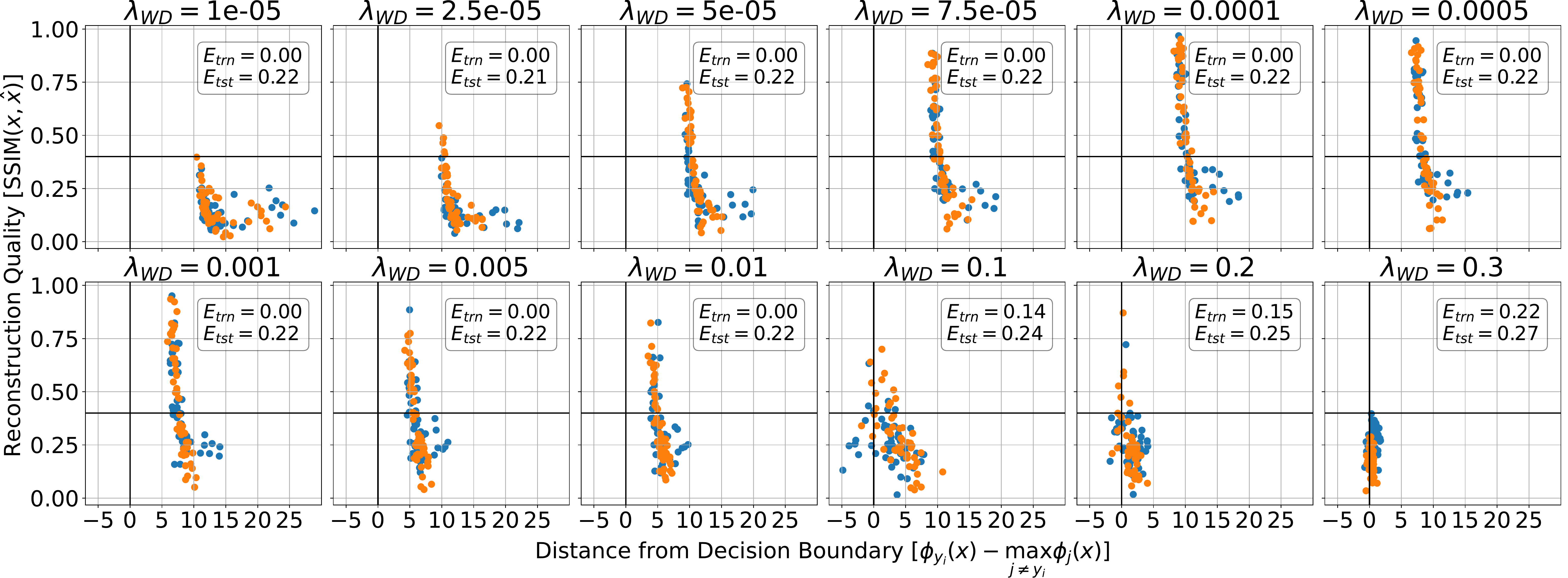}}
\captionof{figure}{\mbox{Weight-Decay ``pushes" more samples to the margin, thus enabling them to be reconstructed}}\label{fig:weight_decay_scatter} 
\end{minipage}

\subsection{Reconstruction From a Larger Number of Samples}

One of the major limitations of \cite{haim2022reconstructing} is that they reconstruct from models that trained on a relatively small number of samples. Specifically, in their largest experiment, a model is trained with only $1000$ samples. Here we take a step further, and apply our reconstruction scheme for a model trained on $5000$ data samples. 

To this end, we trained a 3-layer MLP, where the number of neurons in each hidden layer is $10,000$.  Note that the size of the hidden layer is $10$ times larger than in any other model we used. Increasing the number of neurons seems to be one of the major reasons for which we are able to reconstruct from such large datasets, although we believe it could be done with smaller models, which we leave for future research. 
We used the CIFAR100 dataset, with 50 samples in each class, for a total of $5000$ samples.

\begin{figure}[ht]
\begin{tabular}{c}
     \includegraphics[width=\textwidth]{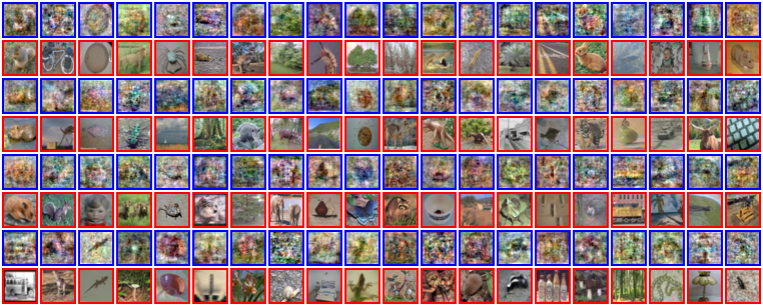} \\
     (a) Full Images. Original samples from the training set (\textcolor{red}{\textit{red}}) and reconstructed results (\textcolor{blue}{\textit{blue}})\\
     \includegraphics[width=\textwidth]{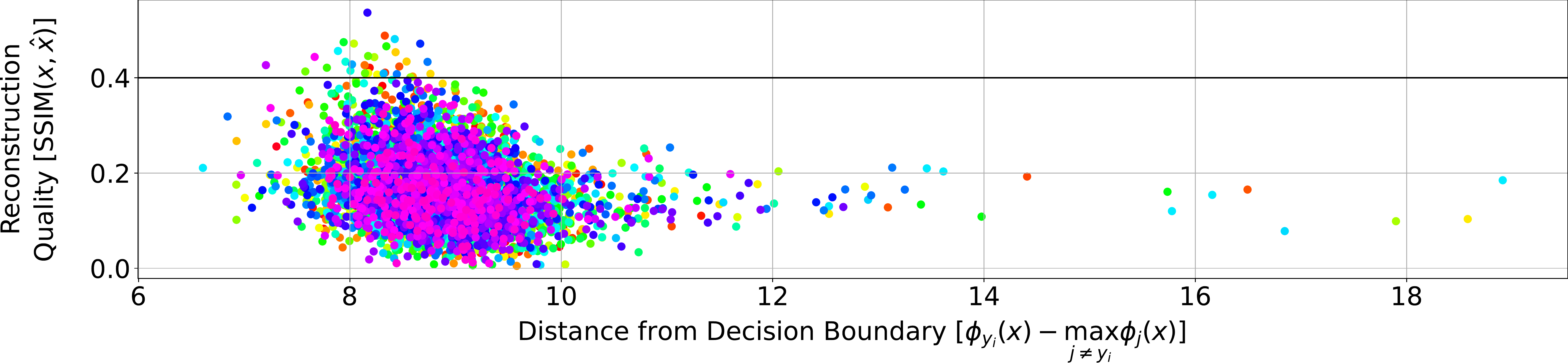} \\
     (b) Scatter plot (similar to \cref{fig:c_vs_dpc}). 
\end{tabular}
\caption{Reconstruction from a model trained on $50$ images per class from the CIFAR100 dataset ($100$ classes, total of $5000$ datapoints)}
\label{fig:cifar_100_scatter}
\centering
\end{figure}

In \cref{fig:cifar_100_scatter}a we give the best reconstructions of the model. Note that although there is a degradation in the quality of the reconstruction w.r.t a model trained on less samples, it is still clear that our scheme can reconstruct some of the training samples to some extent. In \cref{fig:cifar_100_scatter}b we show a scatter plot of the SSIM score w.r.t the distance from the boundary, similar to \cref{fig:c_vs_dpc}a. Although most of the samples are on or close to the margin, only a few dozens achieve an SSIM$>0.4$. This may indicate that there is a potential for much more images to reconstruct, and possibly with better quality.

\vspace{-.6cm}
\section{Conclusions and Future Work}
In this paper we have shown several improvements over \cite{haim2022reconstructing}. Most notably, we have shown that it is possible to reconstruct training data in a multi-class setting, compared to only a binary classification setting in the previous work. Additionally, \cite{haim2022reconstructing} showed reconstructions only from networks trained with small initialization. Here we show reconstructions from networks trained with weight decay, which is much more standard than a small initialization scale. Finally, we show it is possible to reconstruct from models trained on $5000$ data samples, which is 5 times more than the largest trained model in \cite{haim2022reconstructing}

There are a couple of future research directions that we think might be interesting. First, to extend our reconstruction scheme to more practical models such as CNN and ResNets. Second, to reconstruct from models trained on more data, such as the entire CIFAR 10/100 datasets, or different types of data such as text, time-series or tabular data. Finally, it would be interesting to find privacy schemes which could protect from reconstruction attacks by specifically protecting samples which lie on the margin.

\subsubsection*{Acknowledgments}
This project received funding from the European Research Council (ERC) under the European Union’s Horizon 2020 research and innovation programme (grant agreement No 788535), and ERC grant 754705, and from the D. Dan and Betty Kahn Foundation.
GV acknowledges the support of the NSF and the Simons Foundation for the Collaboration on the Theoretical Foundations of Deep Learning.

\bibliography{main}

\begin{thebibliography}{20}
\providecommand{\natexlab}[1]{#1}
\providecommand{\url}[1]{\texttt{#1}}
\expandafter\ifx\csname urlstyle\endcsname\relax
  \providecommand{\doi}[1]{doi: #1}\else
  \providecommand{\doi}{doi: \begingroup \urlstyle{rm}\Url}\fi

\bibitem[Balle et~al.(2022)Balle, Cherubin, and Hayes]{balle2022reconstructing}
Borja Balle, Giovanni Cherubin, and Jamie Hayes.
\newblock Reconstructing training data with informed adversaries.
\newblock \emph{arXiv preprint arXiv:2201.04845}, 2022.

\bibitem[Carlini et~al.(2019)Carlini, Liu, Erlingsson, Kos, and
  Song]{carlini2019secret}
Nicholas Carlini, Chang Liu, {\'U}lfar Erlingsson, Jernej Kos, and Dawn Song.
\newblock The secret sharer: Evaluating and testing unintended memorization in
  neural networks.
\newblock In \emph{28th USENIX Security Symposium (USENIX Security 19)}, pp.\
  267--284, 2019.

\bibitem[Carlini et~al.(2021)Carlini, Tramer, Wallace, Jagielski, Herbert-Voss,
  Lee, Roberts, Brown, Song, Erlingsson, et~al.]{carlini2021extracting}
Nicholas Carlini, Florian Tramer, Eric Wallace, Matthew Jagielski, Ariel
  Herbert-Voss, Katherine Lee, Adam Roberts, Tom Brown, Dawn Song, Ulfar
  Erlingsson, et~al.
\newblock Extracting training data from large language models.
\newblock In \emph{30th USENIX Security Symposium (USENIX Security 21)}, pp.\
  2633--2650, 2021.

\bibitem[Carlini et~al.(2023)Carlini, Hayes, Nasr, Jagielski, Sehwag,
  Tram{\`e}r, Balle, Ippolito, and Wallace]{carlini2023extracting}
Nicholas Carlini, Jamie Hayes, Milad Nasr, Matthew Jagielski, Vikash Sehwag,
  Florian Tram{\`e}r, Borja Balle, Daphne Ippolito, and Eric Wallace.
\newblock Extracting training data from diffusion models.
\newblock \emph{arXiv preprint arXiv:2301.13188}, 2023.

\bibitem[Deng et~al.(2009)Deng, Dong, Socher, Li, Li, and
  Fei-Fei]{deng2009imagenet}
Jia Deng, Wei Dong, Richard Socher, Li-Jia Li, Kai Li, and Li~Fei-Fei.
\newblock Imagenet: A large-scale hierarchical image database.
\newblock In \emph{2009 IEEE conference on computer vision and pattern
  recognition}, pp.\  248--255. Ieee, 2009.

\bibitem[Fredrikson et~al.(2015)Fredrikson, Jha, and
  Ristenpart]{fredrikson2015model}
Matt Fredrikson, Somesh Jha, and Thomas Ristenpart.
\newblock Model inversion attacks that exploit confidence information and basic
  countermeasures.
\newblock In \emph{Proceedings of the 22nd ACM SIGSAC conference on computer
  and communications security}, pp.\  1322--1333, 2015.

\bibitem[Glorot \& Bengio(2010)Glorot and Bengio]{glorot2010understanding}
Xavier Glorot and Yoshua Bengio.
\newblock Understanding the difficulty of training deep feedforward neural
  networks.
\newblock In \emph{Proceedings of the thirteenth international conference on
  artificial intelligence and statistics}, pp.\  249--256. JMLR Workshop and
  Conference Proceedings, 2010.

\bibitem[Haim et~al.(2022)Haim, Vardi, Yehudai, Shamir, and
  Irani]{haim2022reconstructing}
Niv Haim, Gal Vardi, Gilad Yehudai, Ohad Shamir, and Michal Irani.
\newblock Reconstructing training data from trained neural networks.
\newblock \emph{NeurIPS}, 2022.

\bibitem[He et~al.(2015)He, Zhang, Ren, and Sun]{he2015delving}
Kaiming He, Xiangyu Zhang, Shaoqing Ren, and Jian Sun.
\newblock Delving deep into rectifiers: Surpassing human-level performance on
  imagenet classification.
\newblock In \emph{Proceedings of the IEEE international conference on computer
  vision}, pp.\  1026--1034, 2015.

\bibitem[He et~al.(2019)He, Zhang, and Lee]{he2019model}
Zecheng He, Tianwei Zhang, and Ruby~B Lee.
\newblock Model inversion attacks against collaborative inference.
\newblock In \emph{Proceedings of the 35th Annual Computer Security
  Applications Conference}, pp.\  148--162, 2019.

\bibitem[Hitaj et~al.(2017)Hitaj, Ateniese, and Perez-Cruz]{hitaj2017deep}
Briland Hitaj, Giuseppe Ateniese, and Fernando Perez-Cruz.
\newblock Deep models under the gan: information leakage from collaborative
  deep learning.
\newblock In \emph{Proceedings of the 2017 ACM SIGSAC conference on computer
  and communications security}, pp.\  603--618, 2017.

\bibitem[Huang et~al.(2021)Huang, Gupta, Song, Li, and
  Arora]{huang2021evaluating}
Yangsibo Huang, Samyak Gupta, Zhao Song, Kai Li, and Sanjeev Arora.
\newblock Evaluating gradient inversion attacks and defenses in federated
  learning.
\newblock \emph{Advances in Neural Information Processing Systems},
  34:\penalty0 7232--7241, 2021.

\bibitem[Ji \& Telgarsky(2020)Ji and Telgarsky]{ji2020directional}
Ziwei Ji and Matus Telgarsky.
\newblock Directional convergence and alignment in deep learning.
\newblock \emph{Advances in Neural Information Processing Systems},
  33:\penalty0 17176--17186, 2020.

\bibitem[Krizhevsky et~al.(2009)Krizhevsky, Hinton,
  et~al.]{krizhevsky2009learning}
Alex Krizhevsky, Geoffrey Hinton, et~al.
\newblock Learning multiple layers of features from tiny images.
\newblock 2009.

\bibitem[Lyu \& Li(2019)Lyu and Li]{lyu2019gradient}
Kaifeng Lyu and Jian Li.
\newblock Gradient descent maximizes the margin of homogeneous neural networks.
\newblock \emph{arXiv preprint arXiv:1906.05890}, 2019.

\bibitem[Melis et~al.(2019)Melis, Song, De~Cristofaro, and
  Shmatikov]{melis2019exploiting}
Luca Melis, Congzheng Song, Emiliano De~Cristofaro, and Vitaly Shmatikov.
\newblock Exploiting unintended feature leakage in collaborative learning.
\newblock In \emph{2019 IEEE Symposium on Security and Privacy (SP)}, pp.\
  691--706. IEEE, 2019.

\bibitem[Soudry et~al.(2018)Soudry, Hoffer, Nacson, Gunasekar, and
  Srebro]{soudry2018implicit}
Daniel Soudry, Elad Hoffer, Mor~Shpigel Nacson, Suriya Gunasekar, and Nathan
  Srebro.
\newblock The implicit bias of gradient descent on separable data.
\newblock \emph{The Journal of Machine Learning Research}, 19\penalty0
  (1):\penalty0 2822--2878, 2018.

\bibitem[Wang et~al.(2004)Wang, Bovik, Sheikh, and Simoncelli]{wang2004image}
Zhou Wang, Alan~C Bovik, Hamid~R Sheikh, and Eero~P Simoncelli.
\newblock Image quality assessment: from error visibility to structural
  similarity.
\newblock \emph{IEEE transactions on image processing}, 13\penalty0
  (4):\penalty0 600--612, 2004.

\bibitem[Yang et~al.(2019)Yang, Zhang, Chang, and Liang]{yang2019neural}
Ziqi Yang, Jiyi Zhang, Ee-Chien Chang, and Zhenkai Liang.
\newblock Neural network inversion in adversarial setting via background
  knowledge alignment.
\newblock In \emph{Proceedings of the 2019 ACM SIGSAC Conference on Computer
  and Communications Security}, pp.\  225--240, 2019.

\bibitem[Zhang et~al.(2018)Zhang, Isola, Efros, Shechtman, and
  Wang]{zhang2018perceptual}
Richard Zhang, Phillip Isola, Alexei~A Efros, Eli Shechtman, and Oliver Wang.
\newblock The unreasonable effectiveness of deep features as a perceptual
  metric.
\newblock In \emph{CVPR}, 2018.

\end{thebibliography}
\bibliographystyle{iclr2023_conference}

\clearpage

\appendix

\section{Deciding whether a Reconstruction is ``Good"}
\label{sec:ssim_0.4}

\begin{figure}
    \centering
    \begin{tabular}{c}
         \includegraphics[width=\textwidth]{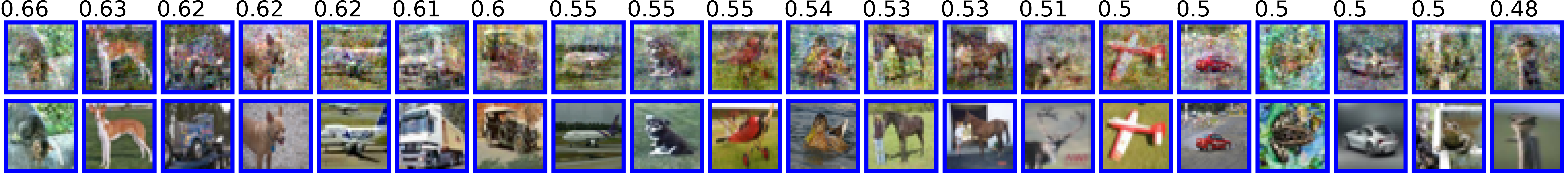}  \\
         \includegraphics[width=\textwidth]{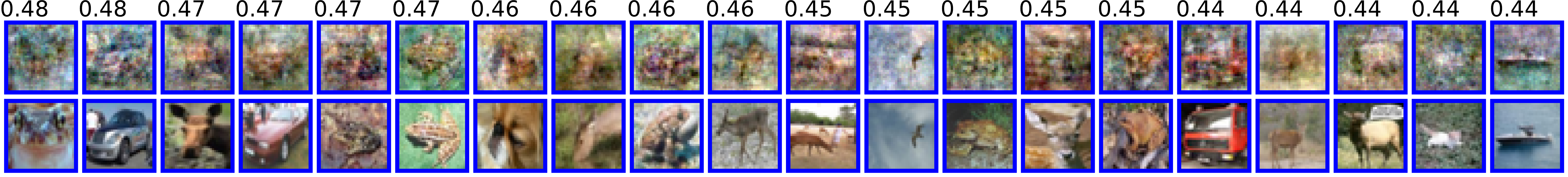}  \\
         \includegraphics[width=\textwidth]{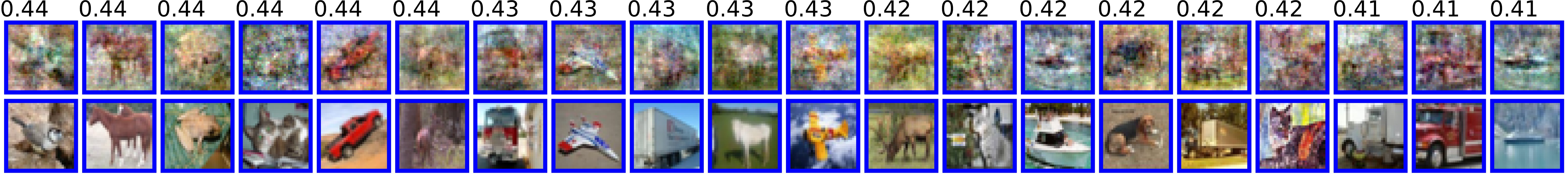}  \\
         \includegraphics[width=\textwidth]{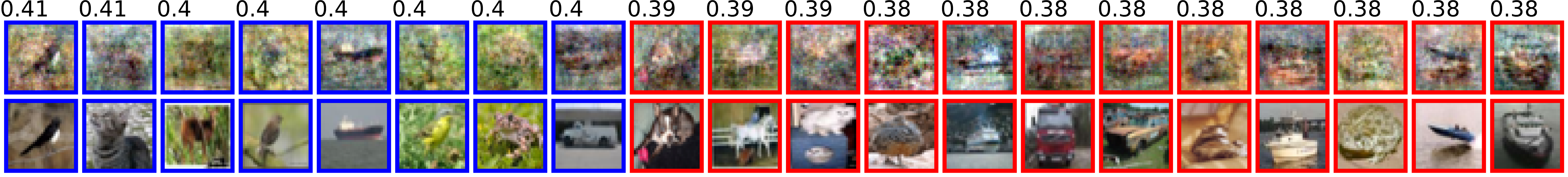}  \\
         \includegraphics[width=\textwidth]{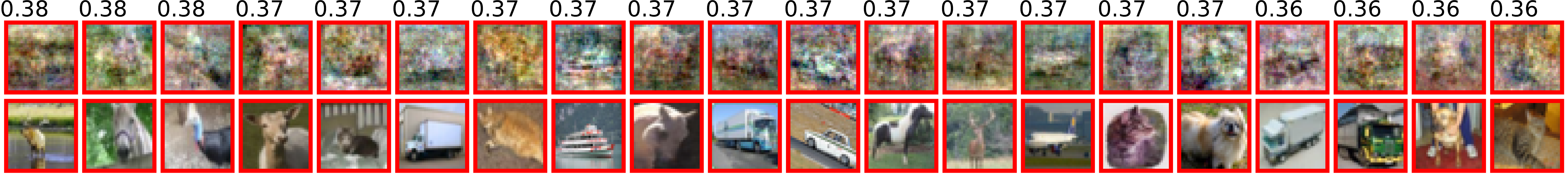}  \\
         \includegraphics[width=\textwidth]{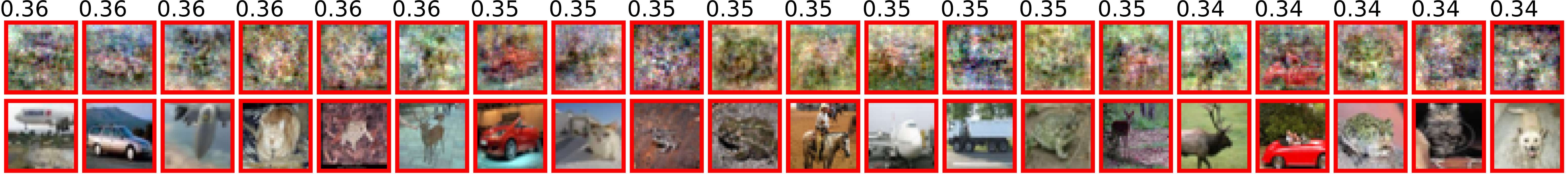} 
    \end{tabular}
    \caption{Justifying the threshold of SSIM$=0.4$ as good rule-of-thumb for a threshold for a ``good" reconstruction. Note that samples with SSIM$>0.4$ (blue) are visually similar. Also some of the samples with SSIM$<0.4$ (red) are similar. In general deciding whether a reconstruction is ``good" is an open question beyond the scope of this paper. The SSIM values are shown above each train-reconstruction pair.}
    \label{fig:ssim_0.4}
\end{figure}

Here we justify our selection for SSIM$=0.4$ as the threshold for what we consider as a ``good" reconstruction. In general, the problem of deciding whether a reconstruction is the correct match to a given sample, or whether a reconstruction is a ``good" reconstruction is equivalent to the problem of comparing between images. No ``synthetic" metric (like SSIM, $l2$ etc.) will be aligned with human perception. A common metric for this purpose is LPIPS~\cite{zhang2018perceptual} that uses a classifier trained on Imagenet~\cite{deng2009imagenet}, but since CIFAR images are much smaller than Imagenet images ($32\times 32$ vs. $224\times 224$) it is not clear that this metric will be better. As a simple rule of thumb, we use SSIM$>0.4$ for deciding that a given reconstruction is ``good". To justify, we plot the best reconstructions (in terms of SSIM) in \cref{fig:ssim_0.4}. Note that almost all samples with SSIM$>0.4$ are also visually similar (for a human). Also note that some of the samples with SSIM$<0.4$ are visually similar, so in this sense we are ``missing" some good reconstructions. In general, determining whether a reconstruction is ``good" is an open question which cannot be dealt in the scope of this paper, and is a future direction for our work.

\section{Experiments with Different Number of Classes and Fixed Training Set Size}
\label{sec:fixed_dataset_size}

\begin{figure}
    \centering
    \includegraphics[width=\textwidth]{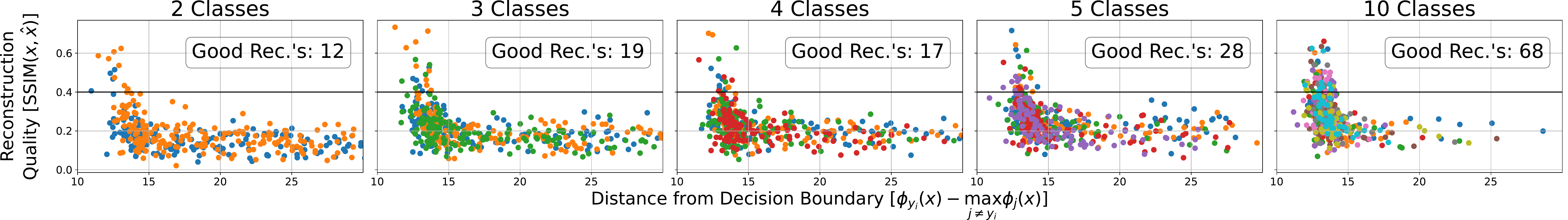}
    \caption{Experiments of reconstruction from models trained on a a fixed training set size ($500$ samples) for different number of classes. Number of ``good" reconstruction is shown for each model.}
    \label{fig:c_vs_dpc_500}
\end{figure}

To complete the experiment shown in \cref{fig:c_vs_dpc}, we also perform experiments on models trained on various number of classes ($C \in \{2,3,4,5,10\}$) but this time with a fixed training set size of $500$ samples (distributed equally between classes). It seems from \cref{fig:c_vs_dpc_500} that the results are not much different from those for $50$ samples per class, and we hypothesize that the model only needs a certain amount of ``support vectors" to support its parameters, and this number is also achieved by $50$ samples per class, and are not harmed if more data is added for each class.
Also note that for models with less classes, not only the number of good reconstruction decreases, but also the quality of reconstruction (lower SSIM). Since we don't have a good heuristic for aggregating the reconstruction quality score of a given model this observation is hard to quantify, but evident from the plot itself.

\end{document}